\documentclass[conference]{IEEEtran}
\IEEEoverridecommandlockouts
\usepackage{cite}
\usepackage{amsmath,amssymb,amsfonts}
\usepackage{algorithmic}
\usepackage{graphicx}

\usepackage{amssymb}

\usepackage{amsmath,amssymb,amsfonts}
\usepackage{algorithmic}
\usepackage{algorithm}
\usepackage{newfloat}
\usepackage{listings}

\usepackage{graphicx} 
\usepackage{courier}
\usepackage{lscape}
\usepackage[hyphens]{url} 

\usepackage{textcomp}
\usepackage{xcolor}
\def\BibTeX{{\rm B\kern-.05em{\sc i\kern-.025em b}\kern-.08em
    T\kern-.1667em\lower.7ex\hbox{E}\kern-.125emX}}

\begin{document}

\title{ Long-only cryptocurrency portfolio management by ranking the assets: a neural network approach \\
}

\author{\IEEEauthorblockN{1\textsuperscript{st} Zijiang Yang}
\IEEEauthorblockA{\textit{Department of Computer Science and Engineering} \\
\textit{New York University}\\
New York City, United States \\
zy3110@nyu.edu}
}

\maketitle


\begin{abstract}
This paper will propose a novel machine learning based portfolio management method in the context of the cryptocurrency market. Previous researchers mainly focus on the prediction of the movement for specific cryptocurrency such as the bitcoin(BTC) and then trade according to the prediction.  In contrast to the previous work that treats the cryptocurrencies independently, this paper manages a group of cryptocurrencies by analyzing the relative relationship. Specifically, in each time step, we utilize the neural network to predict the rank of the future return of the managed cryptocurrencies and place weights accordingly. By incorporating such cross-sectional information,  the proposed methods is shown to profitable based on the backtesting experiments on the real daily cryptocurrency market data from May, 2020 to Nov, 2023. During this 3.5 years, the market experiences the full cycle of bullish, bearish and stagnant market conditions. Despite under such complex market conditions, the proposed method outperforms the existing methods and achieves a Sharpe ratio of 1.01 and annualized return of 64.26\%. Additionally, the proposed method is shown to be robust to the increase of transaction fee.
\end{abstract}


\section{Introduction}
Investment in the financial market is crucial to both retail and institutional investors. Investors hold a number of assets and construct the financial portfolio. In the context of portfolio management, how to allocate the weight to each asset becomes the most crucial and challenging problem. Markowitz, who received the 1990 Nobel price in economics, proposed the foundational theoretical model for modern portfolio theory(MPT) \cite{Markowitz1952}. The main idea behind is to maximize the return while controlling the overall risk. 

Previous researchers have done a lot of work on the traditional financial market, such as the stock markets. However, few have researched on the portfolio problems on the cryptocurrency market. Cryptocurrency is a newly developed digital asset whose value is backed by the cryptographic decentralized technology. The cryptocurrencies’ value on the market rise significantly and the estimated market value of the largest cryptocurrency bitcoin is comparable to gold. While it is gaining popularity around the world, it posts a number of challenges on managing the portfolio of the cryptocurrencies as the prices of the cryptocurrencies move randomly and fluctuate significantly.

Nowadays, more and more researchers have applied machine learning techniques in the financial market. Previous researchers mainly studied the prediction of the price movement of the assets using the historic data \cite{Kimoto1993, Preethi2012,Adebiyi2014,Li2015,Akita2016,Dixon2016}. Then the portfolio can be constructed by allocating more weights on the asset that gives the highest future return. This kind of approach is classical and intuitive  as this problem can be easily formulated as a supervised learning task. 

This paper will focus on constructing the portfolio using neural networks. In contrast to the previous work, this paper will propose a method that forecasts the cross-sectional return instead of the individual return and constructs the portfolio by ranking the assets returns. We compare the performance of the proposed method with the traditional methods and the methods using other machine learning techniques such as random forest, XGboost, SVR and kNN. The result using the neural network is shown to be most profitable, stable and robust under various market conditions including the bullish, bearish and stagnant markets. 

Also, based on the experiments on the recent real trading data, the out-of-sample backtest result has shown that the proposed method outperforms the previous ones. The limitation of the previous methods is that they are suitable to one specific type of market. For example, the trend following methods always allocate more weights on the past winners, hoping that they will have stronger performance in the future. Such strategy may perform poorly in the stagnant market, as the winner keeps reverting back to the average. In contrast, the proposed method focuses on predicting the rank of the return for the cryptocurrency assets for the future time step regardless of the specific market condition and allocates the weight accordingly. By characterizing it as a supervised learning task, it is able to harness the prediction power of the neural network through incorporating both the temporal and cross-sectional information. 

In addition, this paper will discuss the influence of the transaction cost on the portfolio construction and propose viable solution to avoid frequent re-balancing. For simplicity and practicability, this paper will only consider the long-only portfolio, which means short-selling is prohibited.


\section{Related Research}
Previous researchers have done extensive analysis on the traditional algorithms. T.M.
Cover \cite{Cover1991} introduced the notion of universal portfolio that provides a general framework for portfolio construction. In this paper, he summarized several simple and classical algorithms such as buy and hold, constant rebalancing portfolio, universal optimal fixed portfolio. Helmbold et al. \cite{Helmbold1998} extended it and proposed an exponential gradient algorithm that follows the trends. A number of other researchers proposed other types of portfolio. Allan Borodin et al. \cite{Borodin2003} utilized the mean-reversion behaviour and proposed the anti-correlation algorithm. Bin Li et al. \cite{Bin2011} also focused on mean-reversion and generated some algorithms based on it \cite{Bin2011,Bin2012,Bin2013}. 

Recently, some researchers have done a lot of work on the application of machine learning techniques in portfolio management. Fabio D. Freitas et al.\cite{Freitas2009} presented a prediction-based portfolio optimization method using neural network to predict the return and risk. They applied it on the Brazilian market and the experimental results outperformed the traditional mean-variance model.  Ryo Akita et al. \cite{Akita2016} proposed a deep learning method that combines both the numerical and textual information to forecast the  financial time series. The textual information which mainly comes from the newspaper helps improve the prediction accuracy. Jiang, Z. and Liang, J. \cite{Jiang2016} used the reinforcement learning technique on the cryptocurrency market. Fuli Feng et al. \cite{Feng2019} applied the Temporal Graph Convolution(TGC) to model the temporal evolution and the relation network of stocks. Its performance on the NYSE and NASDAQ was shown to be superior to the state-of-the-art stock prediction solutions. 

However, there is limited amount of papers focusing on the cryptocurrency market. The recently developed machine-learning based algorithms are often too complicated to implement. The backtesting period of the experiments is usually very short that does not cover various market conditions.

\section{Methodology}


In this section, we will first review the trading algorithms proposed by the previous researchers and then introduce our proposed machine learning based algorithms. The details of the feature generation and training process will be discussed. At last, we will backtest on the real world data and analyze the out-of-sample results based on the classical performance metrics.

\subsection{Problem Setting}\label{AA}
Suppose we have \(n\) assets in our portfolio. Let \(p_{i,j}\) denotes the price of \(j\)-th asset at time \(i\). The time spans from time 1 to time T. Then the price matrix is given by:
\begin{equation}
\textbf{P}_{1,T} = 
\begin{pmatrix}
p_{1,1} & p_{1,2} & \dots & p_{1,n}\\
p_{2,1} & p_{2,2} & \dots & p_{2,n}\\
\dots \\ 
p_{T,1} & p_{T,2} & \dots & p_{T,n}
\end{pmatrix} 
\end{equation}
Then we define the return vector as 
\begin{equation}
\textbf{r}_t = [\frac{p_{t+1,1}}{p_{t,1}}-1, \frac{p_{t+1,2}}{p_{t,2}}-1, \dots , \frac{p_{t+1,n}}{p_{t,n}}-1 ]^T
\end{equation}
Let  \(w_{i,j}\) denotes the weight of the total wealth invested in \(j\)-th asset at time \(i\) and define weight vector \(\textbf{w}_{t} = [w_{t,1}, w_{t,2}, \dots, w_{t,n}]^T\) as the weight for each asset at time \(t\). So it has the following property: \(\sum_{j=1}^{n} w_{i,j} = 1\). Let \(R_t\) denotes the return of the overall portfolio at time \(t\). It follows:
\begin{equation}
R_t = \textbf{w}_{t}^T \textbf{r}_t
\end{equation}
The goal of the trading algorithm is to compute the $ \textbf{w}_{t}$ given the information up to time $t$ such that the return is maximized while the volatility of this return is minimized.

\subsection{Existing trading algorithms} 
\subsubsection{Benchmark}
\begin{enumerate}
\item BAH 

The simplest trading algorithm is to “buy-and-hold”(BAH) assets. We invest into the assets with the initial weight $\textbf{w}_{1}$ and we never do any rebalancing during the trading period. In our paper, we choose the weight to be uniform, namely $\textbf{w}_{1} = [1/n, 1/n, \dots , 1/n]^T$.

\item  UCRP 

An alternative way to this static BAH algorithm is to dynamically adjust the weight $\textbf{w}_{t}$ during trading period. The simplest example may be using a fixed portfolio weight $ \textbf{w} $ and reinvest the assets according to the this weight $ \textbf{w}$ (T.M. Cover(1991)). We choose the universal constant rebalancing portfolio(UCRP) as our first dynamic trading algorithm. This algorithm sets the weight $\textbf{w}_{t} = [1/n, 1/n, \dots , 1/n]^T$. The difference between UCRP and BAH is that in UCRP the money invested each assets on any day will be equal while in BAH the money may fluctuate with the asset price. 

\item  BCRP(hindsight)

Following the above logic, given the full price matrix, including the past and future information, we are definitely be able to compute a universal optimal fixed portfolio weight $\textbf{w} $ such that the return is maximized \cite{Cover1991}. Obviously, the performance of this portfolio is always at least as good as the best performing individual asset. So we adopt this portfolio(in hindsight) to be a reference as a 'good' portfolio.

\end{enumerate}
\subsubsection{Follow the winner}
The aim of 'follow the winner' strategy is to allocate more weights on the assets that performs better in the past history. 

\begin{enumerate}
\item  UP 

T.M. Cover \cite{Cover1991} introduced the notion of universal portfolio(UP). This algorithm first starts with uniform weight and keeps track of the assets' past history. It then computes the holding value for each asset and assigns the weight proportional to it. Thus, more weight will be allocated on the asset that performs better in the past.

\item  EG

Helmbold et al. \cite{Helmbold1998}  proposed the exponential gradient(EG) portfolio, which tracks the best performing asset so far in exponential return manner. They also added the regularization of the weight to limit the rapid changing of the weight.

\end{enumerate}

\subsubsection{Follow the loser} The aim of 'follow the loser' strategy is to capture the mean reversion behaviors in the assets. It tends to allocate more weights to the assets that perform poorly in the past and hope that these assets can revert back to the mean quickly.

\begin{enumerate}
\item Anti-correlation

This algorithm \cite{Borodin2003} used predictable statistical relations between
all pairs of assets in the market rather than the individual trends. It learned the 'reversal to mean' behaviour adaptively based on the empricial analysis.

\item CWMR/OLMAR/PAMR

Confidence weighted mean reversion(CWMR), online learning moving average reversion(OLMAR) and passive aggressive mean reversion(PAMR) are three types of mean reversion strategies \cite{Bin2011,Bin2012,Bin2013}. They allocate the weights by analying the confidence level of the past return(CWMR), learning from the past moving averages of the assets(OLMAR) and passive aggressive learning of the past return(PAMR). 

\item RMR

Similar to the OLMAR, it makes use of the mean reversion inside the assets. However, previous OLMAR may subject to the noise and outlier. Dingjiang Huang et al. \cite{Huang2016}  adopted a more robust measure($L_{1} $-estimator) to  build up the portfolio.

\end{enumerate}

\subsubsection{Pattern matching} 
This type of algorithms allocate the weights by looking inside the historical price movement and finding the similar price pattern via various proximity measures.

\begin{enumerate}
\item BNN

L. Gyorfi \cite{Udina2006,Walk2008} introduced the kernal-based non-parametric nearest neighbor log-optimal trading algorithm using the kernel based distance measure. Then the algorithm used a nearest neighbor method to find the previous pattern and allocated the weights accordingly.

\item CORN

Similar to the previous BNN, Bin Li et al. \cite{Gopalkrishnan2011} used the correlation-driven non-parametric learning method to build the portfolio.

\end{enumerate}

\subsection{Machine learning based trading algorithms} 
In our paper, we proposed the machine learning based trading algorithms for managing the cryptocurrency portfolio. The algorithms follow the following steps:  
\begin{itemize}
    \item Extract the feature from the information up to time $t$;
    \item Set a prediction target, usually the historical return for each asset; 
    \item Train the machine learning model and use the trained model to compute the weight for time $t$.
\end{itemize}
Let $\textbf{f}_t$ denotes the feature extracted at time $t$ using the previous information and $\textbf{F}_{t-N+1, t}$ denotes the feature matrix for last $N$ days, $\textbf{F}_{t-N+1, t} = [\textbf{f}_{t-N+1}, \textbf{f}_{t-N+2}, \dots, \textbf{f}_{t}]^T$.  Let $\textbf{b}_{t}$ denotes the prediction target and $\textbf{B}_{t-N+1, t}$ denotes the target matrix for last $N$ days, $\textbf{B}_{t-N+1, t} = [\textbf{b}_{t-N+1}, \textbf{b}_{t-N+2}, \dots, \textbf{b}_{t}]^T$. 

\subsubsection{Feature generation}
For the feature generation, we adopt the traditional features that are used by the previous researchers. Specifically, we use the following features:
\begin{itemize}
    \item Last return: the last return for the assets from time $t-1$ to time $t$.
    \item Volatility: the volatility for the assets in training period.
    \item Sharpe ratio: the Sharpe ratio for the assets in training period.
    \item Return rank correlation: it measures how strong of the trend in the training period for individual assets.
    
\end{itemize}

\subsubsection{Transforming the return to rank}

In the previous research, the target we would like to predict is usually the return of assets in the next day, namely $\textbf{b}_t = \textbf{r}_t$. This is pretty intuitive and straightforward. However, predicting the absolute return is a difficult task, especially in such a volatile market. Actually, based on our experimental result in Table.\ref{tabletradingalgoresult}, the approach using the next day return $r_{t}$  as the prediction target may not deliver a satisfactory result.

So, we consider predicting the relative strength of the asset return instead of the absolute return. We transform the next day return $r_{t}$ to the $rank(r_{t})$ as the target we would like to predict. Moreover, we have tried different setups by setting the power of the rank, namely $(rank(r_{t}))^n$ as the prediction target for different $n$. The objective of powering it up is to enlarge the relative return difference between different assets. In our experiment, this $n$ is chosen to be $2, 3, 4$.

For example, suppose we have the information up to time 100. The Table.\ref{tablereturnpredictiontarget} shows the prediction target we are going to predict.

\begin{table}[htb]
\centering
\caption{Prediction target for time 100}
\begin{tabular}{lrrr}
\hline
{ \textbf{Assets}}             & { \textbf{return}} & { \textbf{returnRank}} & { \textbf{returnRank\textasciicircum{}2}} \\ \hline
{ \textbf{LINK  }} & { 0.0863}          & { 10.0}                & { 100.0}                                  \\
{ \textbf{BNB }}  & { -0.0196}         & { 2.0}                 & { 4.0}                                    \\
{ \textbf{EOS  }}  & { -0.0076}         & { 4.0}                 & { 16.0}                                   \\
{ \textbf{ETC  }}  & { -0.0025}         & { 8.0}                 & { 64.0}                                   \\
{ \textbf{BCH  }}  & { -0.0151}         & { 3.0}                 & { 9.0}                                    \\
{ \textbf{TRX  }}  & { 0.0036}          & { 9.0}                 & { 81.0}                                   \\
{ \textbf{ADA  }}  & { -0.0056}         & { 5.0}                 & { 25.0}                                   \\
{ \textbf{XRP  }}  & { -0.0208}         & { 1.0}                 & { 1.0}                                    \\
{ \textbf{BTC  }}  & { -0.0048}         & { 6.0}                 & { 36.0}                                   \\
{ \textbf{ETH  }}  & { -0.0047}         & { 7.0}                 & { 49.0}                                   \\ \hline

\label{tablereturnpredictiontarget}
\end{tabular}

\end{table}

\subsubsection{Machine learning algorithms}

To predict the future return rank, we adopt the machine learning approach by characterising the problem  as a supervised learning problem. Based on the previous research, various machine learning algorithms can be applied to do this task.

\begin{enumerate}

\item k-Nearest-Neighbor

The most straightforward way of making such prediction is to find the similar pattern in the past. So, we consider training the k-Nearest-Neighbor algorithm on the training data and assign the values to the test data based on the similarity. In our experiment, the k is set to be 15 and the distance measure is Euclidean distance.

\item Neural Network

It is a popular machine learning method that mimics the functionality of human brains. It uses a number of interconnected neurons that sends activation signals from one to another. In our algorithm, we adopt the simplest form, multi-layer perceptron. The reason for using it is that it is fully connected. At each time step, the changes across different assets will be related and influenced. In addition to the temporal information for individual asset, more cross-sectional information across the assets will be included and learned. In our experiment, we set the hidden layers size to be (20, 20) and the initial seed is 10.


\item Random Forest

Random forest uses bagging technique to aggregate a number 
of decision trees in parallel from the bootstrap samples of the 
data set. The final decision is the voting result of all the 
decision trees.

\item XGBoost
It uses the idea of extreme gradient boosting. Similar as the random forest, it is a ensemble method that aggregates the decision trees. The difference is that it adds a regularization term and more sophisticated optimization method with the extreme gradient.

\end{enumerate}

\begin{algorithm}
\caption{Multi-layer Perceptron(MLP) Trading Algorithm}\label{alg:cap}
\begin{algorithmic}[1]
\STATE Initialize the MLP parameters by setting the number of extracted features as input dimension and the number of assets as output dimension
\STATE  Set the time for start($t_{trading}$) and end($t_{end}$) of trading
\STATE  Initialize feature lookback window $N \gets 80$
\STATE  Set the model parameter update frequence $m \gets 10$
\STATE   $\textbf{R} \gets [ ]$   

\STATE $t \gets t_{trading}$  
\WHILE{$t <= t_{end}$}    
\STATE Prepare the price matrix $\textbf{P}_{1, t}$
\IF{$(t-t_{trading}) \% m == 0$}
\STATE Compute the feature matrix $\textbf{F}_{t-t_{N}, t-1}$
\STATE Compute the target matrix $\textbf{B}_{t-t_{N}, t-1}$
\STATE Fit the model by minimizing 
\STATE $\left\| model(\textbf{F}_{t-t_{N}, t-1})-\textbf{B}_{t-t_{N}, t-1} \right\|^2_2$
\ENDIF

\STATE Compute the feature vector for time $t$: $\textbf{f}_{t}$
\STATE Compute the weight vector for time $t$: 
\STATE $\textbf{w}_{t} = model(\textbf{f}_{t})$
\STATE Normalize the weight vector for time $t$: 
\STATE $\textbf{w}_{t} = \textbf{w}_{t}/sum(\textbf{w}_{t})$
\STATE The portfolio return for time $t$ is given by:
\STATE $R_{t} = \textbf{w}_{t}^{T}\textbf{r}_{t}$
\STATE Append $R_{t}$ to $\textbf{R}$ 
\STATE $t \gets t+1 $  
\ENDWHILE
\end{algorithmic}
\end{algorithm}

\subsection{Adding the decay}
Based on the preliminary result, the weight vector computed via the proposed method changes frequently across the trading period. As a result, it would produce a large amount of commission fee and reduce a large amount of profit. Thus, we introduce the decay to the weight. We define the new weight \(\widetilde{w}_t\) as weighted sum of the past weights and the current predicted weight \(\widehat{w}_{t}\):
\begin{equation}
\widetilde{w}_t = (\sum_{i=1}^{l}\alpha^i\widetilde{w}_{t-i}  + \widehat{w}_{t})/(1+\sum_{i=1}^{l}\alpha^i)
\end{equation}
where \(\alpha\) is the decay rate and \(l\) is the decay length. In our experiment, we choose the \(\alpha\) to be 0.7 and the \(l\) to be 1.

\subsection{Evaluation Metrics}
To analyze the performance of the trading algorithms above, we measure the out-of-sample portfolio performance via the following metrics. 
\begin{enumerate}
\item Annualized Return and Volatility \\
Firstly, we are concerned about the profitability, we measure the annualized return: 
$\bar{R} = \sqrt{250}\sum_{t=1}^{T}R_t $.
For measuring the stability of the return, we compute the annualized volatility: $\sigma_R = \sqrt{250}\sqrt{\frac{1}{T}\sum_{t=1}^{T}(R_t - \bar{R})^2}$ 
\item Winning Percentage \\
We basically just count the percentage of the days that generate positive return.
\item Sharpe Ratio \\
It is also a traditional metric for measuring stability of the return. As is one of the most widely used methods for measuring risk-adjusted relative returns, it computes the ratio of annualized return over annualized volatility: $SR = \bar{R}  / \sigma_R $

\item Profit Factor \\
It is another metric for measuring the profitability of the trading algorithms. It is the ratio of profit to loss in the designated trading periods. Higher this ratio is, the more profitable the algorithm will be.
\begin{equation}
    PF = \sum_{t=1}^{T}R_t\textbf{1}[R_t >= 0] / \sum_{t=1}^{T}R_t\textbf{1}[R_t < 0]
\end{equation}
where $\textbf{1}[\dots]$ is the indicator function.

\item Information Ratio \\
This is a measurement of portfolio returns beyond the returns of a benchmark, usually an index, divided by the volatility of those returns. It evaluates how well the algorithm achieves higher return in excess of the benchmark, given the risk taken.
Let
\begin{equation}
    \delta R_t = R_t - R_{benchmark, t} 
\end{equation}
Then
\begin{equation}
    IR = E[\delta R_t] / \sigma_{\delta R_t}
\end{equation}

\item Maximum Drawdown \\
The maximum drawdown is a measurement of the loss of the portfolio value from the peak to the trough. It is measuring the downside risk, indicating the largest possible loss this algorithm can incur over the trading period.

\end{enumerate}

\section{Experiments and Results}

\subsection{Dataset}
We downloaded the cryptocurrency historical daily data from the CoinGecko website. \footnote{The link is for bitcoin historical data: \url{https://www.coingecko.com/en/coins/bitcoin/historical_data#panel}}  We extracted the data for 10 coins with largest market capital as of 2023/11/01. The daily data ranges from 2020/05/01 to 2023/11/01. The following table lists out the coins we selected:

\begin{center}
\begin{tabular}{ |c|c|c|c|c| } 
 \hline
 BTC & ETH & EOS & ETC & BCH  \\ 
 \hline
 TRX & ADA & XRP & LINK & BNB \\  
 \hline
\end{tabular}
\end{center}

\begin{figure}[htbp]
\centerline{\includegraphics[width=\linewidth]{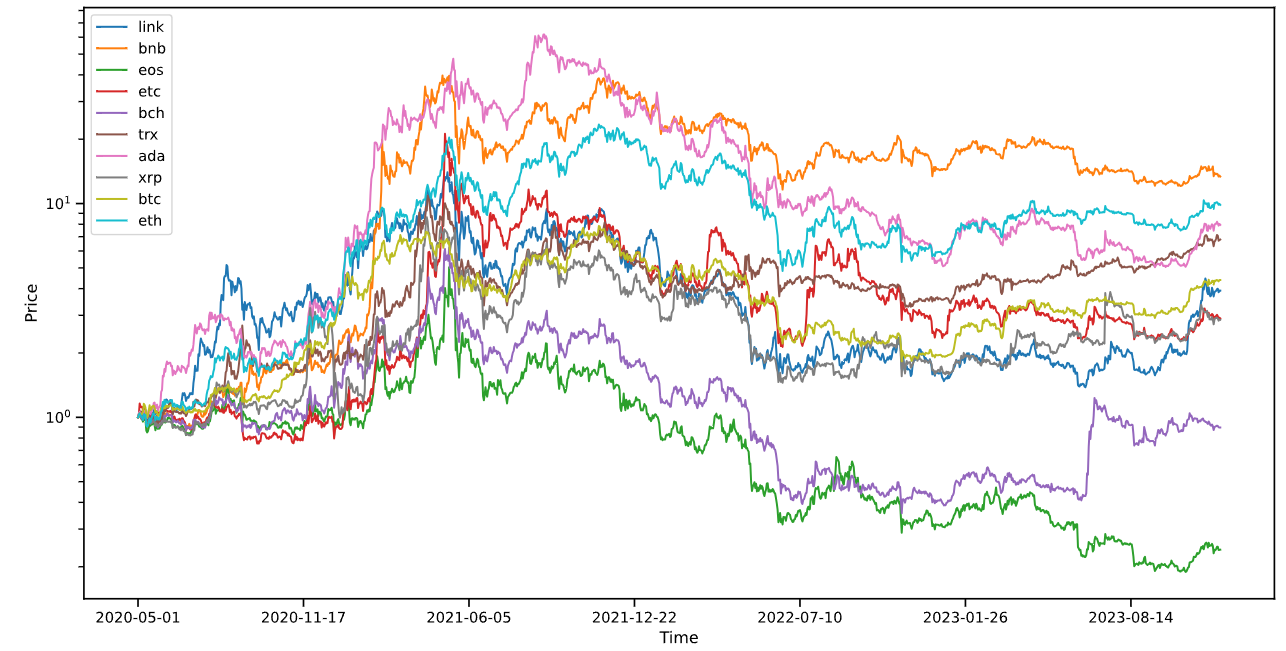}}
\caption{Price series for the coins(normalized to 1 at the beginning; use log scale on y axis)}
\label{figprice}
\end{figure}

We plot out the price series for each coin from 2020/05/01 to 2023/11/01 in the Figure.\ref{figprice}. The prices are normalized to 1 at the date of 2020/05/01 and the y axis is in logarithmic scale. During this 3.5 years period, the cryptocurrency market experiences full cycle of all the market conditions: (1)bullish market(2020/05/01 to 2021/07/01); (2)bearish market(2021/07/01 to 2022/07/01); (3)stagnant market(2022/07/01 to 2023/11/01). 
 
 Moreover, in order to have a better picture of the movement of the coin prices, the daily return statistics is summarized in the Table. \ref{tabledailyreturn}. We can see that volatility of the daily return is very large, reaching 5\%. In extreme cases, the coin price can ramp up by 70\% or drop by 30\% in one day. 
 
\begin{figure}[htbp]
\centerline{\includegraphics[width=\linewidth]{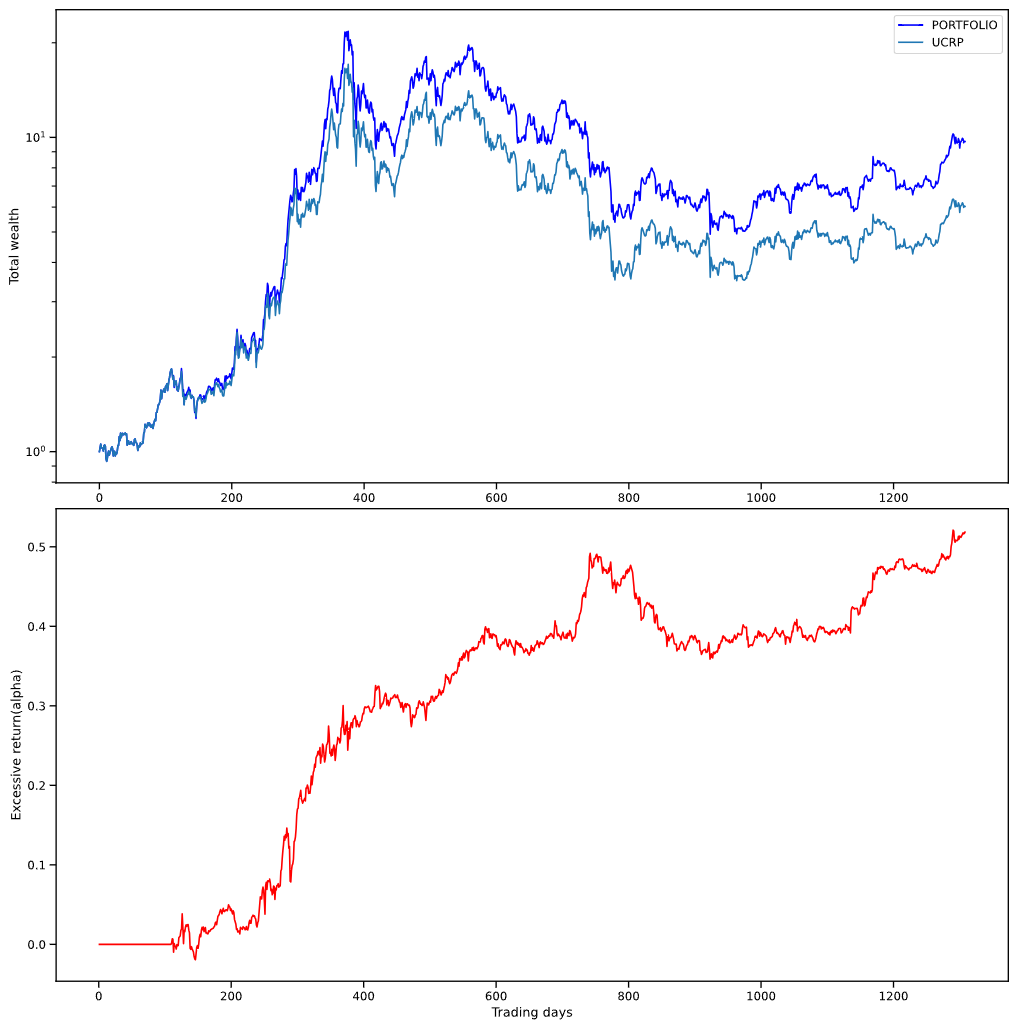}}
\caption{Cumulative wealth of the multi-layer perceptron(MLP) trading algorithm and its excessive return(alpha) with respect to the UCRP(benchmark)}
\label{figportalpha}
\end{figure}

\begin{table*}[htb]
\centering 
\caption{Daily return statistics for cryptocurrencies from 2020/05/01 to 2023/11/01 }
\begin{tabular}{lrrrrrrrrrr }
\hline
Statistics &  LINK        &    BNB  & EOS   & ETC  & BCH  & TRX  & ADA  & XRP  & BTC  & ETH      \\ \hline
count       & 1309        & 1309        & 1309        & 1309        & 1309        & 1309        & 1309        & 1309        & 1309    & 1309    \\
mean        & 0.0027      & 0.0032      & 0.0004      & 0.0024      & 0.0012      & 0.0025      & 0.0029      & 0.0025      & 0.0017  & 0.0027  \\
std         & 0.0575      & 0.0502      & 0.0542      & 0.0583      & 0.0519      & 0.0455      & 0.0518      & 0.0608      & 0.0331  & 0.0433  \\
min         & -0.3565     & -0.3188     & -0.3777     & -0.3255     & -0.3431     & -0.3140     & -0.2461     & -0.4228     & -0.1585 & -0.2630 \\
25\%        & -0.0279     & -0.0153     & -0.0211     & -0.0222     & -0.0221     & -0.0155     & -0.0238     & -0.0206     & -0.0125 & -0.0175 \\
50\%        & 0.0028      & 0.0017      & 0.0001      & -0.0000     & 0.0010      & 0.0022      & 0.0006      & 0.0004      & 0.0006  & 0.0014  \\
75\%        & 0.0311      & 0.0213      & 0.0233      & 0.0223      & 0.0225      & 0.0189      & 0.0235      & 0.0208      & 0.0159  & 0.0241  \\
max         & 0.3038      & 0.7379      & 0.5258      & 0.4467      & 0.5267      & 0.3967      & 0.3089      & 0.7237      & 0.1925  & 0.2453 \\
\hline
\end{tabular}
\label{tabledailyreturn}
\end{table*}

\begin{table*}[htb]
\centering
\caption{Performance of traditional trading strategies on cryptocurrency data from 2020/05/01 to 2023/11/01}
\begin{tabular}{lrrrrrrr}
\hline 
Algorithm & Profit & Sharpe & Information ratio & Annualized  & Maximum  & Winning  & Annualized  \\

& Factor & & w.r.t. UCRP & Return(\%) & Drawdown(\%) & Percentage(\%) & Volatility(\%) \\ \hline
\textit{Benchmark}\\  
BAH       & 1.10   & 0.83   & -0.21       & 52.82              & 77.38        & 54.20        & 63.83                  \\
UCRP       & 1.10   & 0.86   & 0.00        & 55.89              & 79.53           & 55.05        & 64.83                  \\
BCRP(hindsight)      & 1.16   & 1.11   & 0.61        & 75.65              & 71.22           & 54.43        & 68.40                  \\  \hline
\textit{Follow the winner} \\ 
UP        & 1.10   & 0.86   & -0.04       & 55.84              & 79.14           & 55.12        & 64.63                  \\
EG        & 1.10   & 0.86   & -0.01       & 55.88              & 79.31           & 55.05        & 64.70                  \\ \hline
\textit{Follow the loser} \\  
Anti-correlation   & 1.00   & 0.35   & -0.67       & 26.31              & 88.83           & 51.22        & 74.63                  \\
CWMR      & 0.89   & -0.20  & -1.67       & -16.13             & 95.52           & 50.69        & 79.10                  \\
OLMAR     & 0.92   & 0.00   & -1.22       & 0.19               & 95.66           & 51.22        & 82.40                  \\
PAMR      & 0.89   & -0.20  & -1.67       & -16.01             & 95.65           & 50.76        & 79.10                  \\
RMR       & 0.93   & 0.07   & -0.96       & 6.28               & 96.15           & 51.61        & 87.37                  \\  \hline
\textit{Pattern matching} \\  
BNN       & 1.06   & 0.70   & 0.03        & 57.80              & 83.71           & 51.68        & 82.97                  \\
CORN      & 1.05   & 0.63   & -0.07       & 52.29              & 84.04           & 51.20        & 82.93                  \\
 \hline 
\end{tabular}
\label{tabletraditional}
\end{table*}

\begin{table*}[htb]
\centering
\caption{Performance of machine learning trading strategies on cryptocurrency data from 2020/05/01 to 2023/11/01}
\begin{tabular}{lrrrrrrr}
\hline  
Algorithm & Profit & Sharpe & Information ratio & Annualized  & Maximum  & Winning  & Annualized  \\

& Factor  & & w.r.t. UCRP & return(\%) & Drawdown(\%) & pct(\%) & volatility(\%) \\ \hline
MLP($return$)  & 1.10   & 0.87   & 0.01        & 55.95              & \textbf{76.67}          & 54.66        & 64.32                  \\
MLP($returnRank$)  & 1.11   & 0.90   & 0.22        & 57.81              & 79.61           & 54.89        & 63.99                  \\
MLP($returnRank^2$)  & \textbf{1.13}   & \textbf{1.01}   & \textbf{0.94}        & \textbf{64.26}              & 77.40           & 55.05        & \textbf{63.58}                  \\
MLP($returnRank^3$)  & 1.11   & 0.91   & 0.26        & 58.20              & 78.72           & 54.97        & 63.90                  \\
MLP($returnRank^4$)  & 1.10   & 0.87   & -0.08       & 55.19              & 79.98           & 55.20        & 63.71                  \\
XGboost($return$)  & 1.10   & 0.87   & 0.20        & 57.69              & 81.90           & 55.28       & 66.01                  \\
XGboost($returnRank$)  & 1.11   & 0.90   & 0.32        & 58.75              & 80.16           & 54.97        & 65.18                  \\
XGboost($returnRank^2$)  & 1.09   & 0.79   & -0.60       & 50.67              & 82.22           & 54.20        & 64.42                  \\
XGboost($returnRank^3$)  & 1.09   & 0.81   & -0.36       & 52.72              & 81.05           & 54.59        & 64.74                  \\
XGboost($returnRank^4$) & 1.10   & 0.87   & 0.07        & 56.54              & 79.16           & 54.66        & 65.28                 \\ 
RandomForest($return$) & 1.11   & 0.91   & 0.38        & 59.46              & 81.09           & 54.97        & 65.64                  \\ 
RandomForest($returnRank$)  & 1.11   & 0.92   & 0.29        & 58.62              & 79.90           & \textbf{55.58}        & 63.69                  \\  
RandomForest($returnRank^2$)  & 1.11   & 0.88   & 0.06        & 56.46              & 81.16           & 55.50        & 63.96                  \\  
RandomForest($returnRank^3$)  & 1.11   & 0.90   & 0.25        & 58.20              & 80.13           & 54.97        & 64.44                  \\  
RandomForest($returnRank^4$) & 1.10   & 0.86   & -0.03       & 55.58              & 81.52           & 55.05        & 64.41                  \\  
kNN($return$)  & 1.11   & 0.89   & 0.27        & 58.23              & 81.78          & 54.43        & 65.24                  \\  
kNN($returnRank$)  & 1.12   & 0.94   & 0.51        & 60.42              & 79.39           & 54.43        & 64.36                  \\  
kNN($returnRank^2$)  & 1.11   & 0.91   & 0.31        & 58.66              & 79.25           & 54.66        & 64.32                  \\  
kNN($returnRank^3$)  & 1.11   & 0.89   & 0.19        & 57.60              & 79.84           & 55.20        & 64.53                  \\ 
kNN($returnRank^4$) & 1.10   & 0.86   & -0.01       & 55.77              & 80.46           & 55.20        & 64.56                  \\ \hline  
\label{tabletradingalgoresult}
\end{tabular}
\end{table*}

\begin{table*}[htb]
\centering  
\caption{Performance of MLP($returnRank^2$) trading strategies on cryptocurrency data under different commission fee from 2020/05/01 to 2023/11/01}
\begin{tabular}{lrrrrrrr}
\hline  
Fee & Profit & Sharpe & Information ratio & Annualized  & Maximum  & Winning  & Annualized  \\

& Factor  & & w.r.t. UCRP & return(\%) & Drawdown(\%) & pct(\%) & volatility(\%) \\ \hline
0.000   & 1.13        & 1.01               & 0.94                   & 64.26       & 77.40 & 55.05 & 63.58        \\
0.00025  & 1.13        & 0.99               & 0.81                   & 63.06       & 78.04 & 55.05 & 63.59        \\
0.0005  & 1.12        & 0.97               & 0.67                   & 61.86       & 78.71 & 54.97 & 63.59        \\
0.00075  & 1.12        & 0.95               & 0.54                   & 60.65       & 79.37 & 54.97 & 63.59        \\
0.001 & 1.12        & 0.93               & 0.40                   & 59.45       & 80.00 & 54.97 & 63.59        \\
0.00125  & 1.11        & 0.92               & 0.27                   & 58.25       & 80.61 & 54.97 & 63.60        \\
0.0015  & 1.11        & 0.90               & 0.13                   & 57.05       & 81.21 & 54.82 & 63.60                     \\ \hline
\end{tabular} 
\label{tablefeetest} 
\end{table*}



\subsection{Trading result} 
We apply the trading strategies on the cryptocurrency data from 2020/05/01 to 2023/11/01. The weight we assign are positive for all assets, which means it is a long only portfolio. For simplicity, we first generate the result with no transaction fee. The results are shown in the Table.\ref{tabletraditional} and Table.\ref{tabletradingalgoresult}.

\subsection{Discussion on the trading result} 

\subsubsection{Traditional strategy trading result}
We first look at the trading result for traditional trading strategies. The results are tabulated in the Table.\ref{tabletraditional}. The first three rows show the performance of the benchmark strategies. As the cryptocurrency experience a rather bullish market in the trading period, the simple BAH strategy gives a decent result with Sharpe ratio 0.83. The universal constant rebalancing portfolio(UCRP) with the uniform weight is set to be the benchmark. The information ratios of the other strategies are computed with respect to UCRP. Based on the table, it is reported that BAH has a negative information ratio, indicating it cannot beat the UCRP benchmark. The BCRP, which chooses the best constant weight in hindsight, will definitely beat the UCRP, achieving the information ratio of 0.61.

In the middle rows, we have the result of 'follow-the-winner' strategies. The performance of UP and EG is satisfactory with Sharpe ratio of 0.86. As the market is bullish, tracking the winner may generate a decent profit. However, compared to the benchmark CRP, it is  not good enough. The information ratio is negative for these two strategies.

Also, we have the results of 'follow-the-loser' strategies. Unfortunately, the performance of these strategies are pretty bad with very low Sharpe ratios. Some strategies even generate the negative returns in such a bullish market. The reason behind is that some coins keeps dropping in the trading period. For instance, the coin EOS drops from 1 to 0.4, which is a substantial drop,  during this period. As the nature of these strategies is betting on mean reversion, the strategies keep tracks of these 'loser' coins and place weights on them hoping the price may revert back. In the cryptocurrency market, the price of the coins may go to zero. For the management of  the cryptocurrency portfolio, We would not recommend the 'follow-the loser' strategies, as they are too risky.
Finally, we have the 'pattern matching' strategies. The strategies look for the similar pattern in the historical price patterns and assign the weight accordingly. The performance of these strategies are acceptable with similar Sharpe ratio as the 'follow-the-winner' strategies. However, the information ratios for them are still very low.

To sum up, the performance of the traditional trading strategies are acceptable. They are able to deliver positive annualized returns and Sharpe ratios in such complex market conditions. However, they cannot outperform even the simplest uniform constant weight portfolio. Note that the results shown in the Table.\ref{tabletraditional} are without any transaction fee. The results can be even worse with the consideration of the transaction cost. 

\subsubsection{Proposed machine learning strategy trading result}
Let us analyze the trading result for the proposed machine learning strategies. The results are presented in the Table.\ref{tabletradingalgoresult}. Generally speaking, all the strategies achieve decent performance. All the Sharpe ratios are greater than 0.79 and all the annualized returns are greater than 50\%. In addition, almost all the strategies achieve the positive information ratio with respect to the benchmark UCRP. The cummulative wealth of  the best MLP based trading algorithm and its excessive alpha over the benchmark is plotted in the Figure.\ref{figportalpha}. We can clear see that it achieves a stable excessive return compared to the benchmark UCRP. It consistently outperforms the benchmark in the first 800 trading days that includes both the bullish and bearish markets. Then it drops at the end of the bearish market. In the last 300 trading days, it continues to beat the benchmark during the stagnant market.

\subsection{Sensitivity analysis of the transaction fee} 
The previous experimental results are built under the zero transaction fee condition. To investigate the impact of the transaction fee on the proposed trading algorithm, we gradually increase the fee level and compute the final trading results. The transaction fee ranges from 0\% to 0.15\%. Actually, 0.15\% is considerably large compared to the transaction fee in the mainstream centralized trading exchanges such OKX and Binance. The result is tabulated in the Table.\ref{tablefeetest}. It is reported that the best performer MLP($returnRank^2$) achieves positive information ratio with respect to the benchmark UCRP under various transaction fee levels. The proposed algorithm's profit is robust to the transaction fee.

\section{Conclusion and Future Work}
In this paper, we propose a novel machine learning based portfolio management algorithm. In stead of treating each asset independently, this algorithm takes the cross-sectional relations into account. By predicting the rank of the assets in each time step, we have shown that this algorithm is practical and profitable. Based on the experimental data on the cryptocurrency market that span the bullish, bearish and stagnant cycles, it is reported that the proposed method outperforms the existing trading algorithm. In the future, we will investigate how to generate more effective input features and apply the proposed method on the traditional market.



\bibliographystyle{elsarticle-num-names} 


\end{document}